\documentclass{article}

\usepackage{spconf,amsmath,graphicx}
\usepackage{cite}
\usepackage{amsmath,amssymb,amsfonts}
\usepackage{algorithmic}
\usepackage{graphicx}
\usepackage{textcomp}
\usepackage{xcolor}
\usepackage{hyperref}
\usepackage{bbm}
\usepackage{multirow}

\usepackage[compact,center]{titlesec}
\titleformat*{\section}{\centering\large\scshape\bfseries}
\titleformat*{\subsection}{\bfseries}
\titlelabel{\thetitle.\quad}
\AtBeginDocument{
  \setlength\abovedisplayskip{0pt}
  \setlength\belowdisplayskip{0pt}}
\usepackage[margin=0pt, font=small]{caption}
\usepackage[activate={true,nocompatibility},final,tracking=false,kerning=true,spacing=true,factor=1100,stretch=10,shrink=20]{microtype}
\usepackage[moderate]{savetrees}
\title{NEV-NCD: Negative Learning, Entropy, and Variance Regularization \\ based Novel Action Categories Discovery}
\name{\begin{tabular}{c}Zahid Hasan\textsuperscript{*}, Masud Ahmed\textsuperscript{*}, Abu Zaher Md Faridee\textsuperscript{+}, Sanjay Purushotham\textsuperscript{*} \\Heesung Kwon\textsuperscript{\$}, Hyungtae Lee\textsuperscript{\$}, Nirmalya Roy\textsuperscript{*}\end{tabular}
\thanks{This research is supported by the NSF CAREER grant  \#1750936, REU Site grant \#2050999 and U.S. Army grant \#W911NF2120076.}
}

\address{
\textsuperscript{*}Department of Information Systems, University of Maryland, Baltimore County, USA\\
\textsuperscript{+}Amazon, USA;
\textsuperscript{\$}DEVCOM Army Research Laboratory (ARL), USA}
\begin{document}
\maketitle
\begin{abstract}

\textit{Novel Categories Discovery} (NCD) facilitates learning from a partially annotated label space and enables deep learning (DL) models to operate in an open-world setting by identifying and differentiating instances of novel classes based on the labeled data notions. One of the primary assumptions of NCD is that the novel label space is perfectly disjoint and can be equipartitioned, but it is rarely realized by most NCD approaches in practice. To better align with this assumption, we propose a novel single-stage joint optimization-based NCD method, \textbf{N}egative learning, \textbf{E}ntropy, and \textbf{V}ariance regularization NCD (\textbf{NEV}-NCD). We demonstrate the efficacy of NEV-NCD in previously unexplored NCD applications of video action recognition (VAR) with the public UCF101 dataset and a curated in-house partial action-space annotated multi-view video dataset. We perform a thorough ablation study by varying the composition of final joint loss and associated hyper-parameters. During our experiments with UCF101 and multi-view action dataset, NEV-NCD achieves $\approx~83\%$ classification accuracy in test instances of labeled data. NEV-NCD achieves $\approx~70\%$ clustering accuracy over unlabeled data outperforming both naive baselines (by $\approx~40\%$) and state-of-the-art pseudo-labeling-based approaches (by $\approx~3.5\%$) over both datasets. Further, we propose to incorporate optional view-invariant feature learning with the multiview dataset to identify novel categories from novel viewpoints. Our additional view-invariance constraint improves the discriminative accuracy for both known and unknown categories by $\approx~10\%$ for novel viewpoints.

\end{abstract}


\section{Introduction}

Deep learning (DL) based approaches form the backbone of modern video understanding, video action recognition (VAR), surveillance, and human-computer interactions. 
They learn to extract relevant spatiotemporal action features from videos and map them to predefined action classes.
The traditional VAR models often assume the existence of exact labels (often laboriously annotated) for all action classes and fail to comprehend novel unlabeled action categories. 
Besides these models often fail to identify known classes from different viewpoints due to a lack of view-dependent features learning.

\begin{figure}[]
 \centering
 \includegraphics[width=\linewidth]{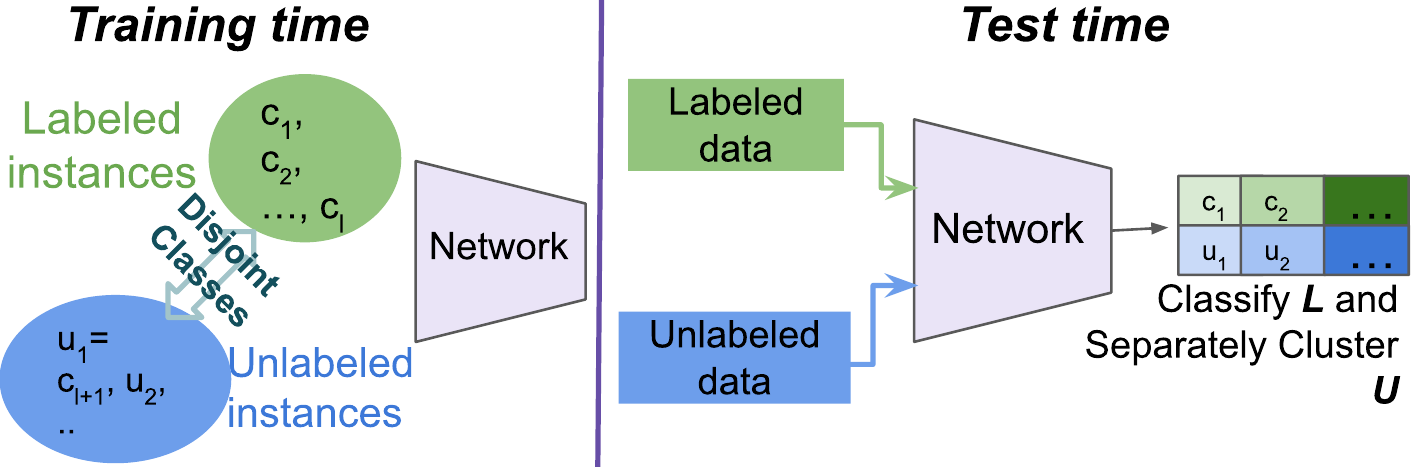}
 \caption{NCD objective aims to detect and structure the unlabeled novel categories based on the semantics of disjoint labeled data.}
 \label{fig:ncd_objective}
\end{figure}

\textit{Novel Categories Discovery} (NCD)~\cite{han2019learning, han2021autonovel, han2020automatically}, an emerging field of DL, aims at more generalized model development under the partial class-space labeled data and alleviates the need for class-specific exact labels for all classes. NCD methods utilize the semantics knowledge of class notions from the labeled data to detect and structure unlabeled novel classes while retaining the classification performance (Figure~\ref{fig:ncd_objective}). In this research, we study a novel NCD method in VAR to extract generalized spatial-temporal action features from videos based on labeled class notions to identify and cluster unlabeled actions.

The current NCD approaches rely on either a two-stage or a single-stage joint optimization \cite{joseph2022spacing} with various data-specific assumptions (simultaneous/sequential data availability, disjoint classes (mutually exclusive sets) between labeled and unlabeled data, knowledge about novel class quantity and their uniform distribution). These approaches optimize the classification loss of labeled data and cluster unlabeled data based on the class notions and semantics of the labeled data. The clustering constraints for unlabeled data are the primary key to NCD and can be enforced by pairwise similarity losses \cite{hsu2017learning}, pseudo-labeling \cite{jia2021joint}, prototype learning \cite{zhang2022automatically}, Sinkhorn-Knopp clustering \cite{fini2021unified}, and self-training \cite{asano2019self}, using the uniformity property.


These NCD clustering objectives aim to achieve low entropy and equal variance among the unlabeled classes. However, these approaches fail to explicitly capitalize on the NCD's assumption of unique, uniform, and disjoint labels for the unlabeled instances.  
Hence, we formulate the NCD's disjoint properties as learning from complementary labels for unlabeled data and find the similarities of fair $k-$face dice statistics with the unique and uniform conditions of the unlabeled data. We propose an intuitive solution to achieve low-entropy, equal class variance aligning equipartition constraint for the novel classes in separate embedding spaces from the labeled classes taking inspiration from negative learning~\cite{kim2019nlnl} and VICreg~\cite{bardes2021vicreg}. We hypothesize that negative learning (disjoint labels), explicit entropy (unique label), and variance (equipartition) regularization would cluster the unlabeled instances aligning with the semantics of the partially labeled data. 


Further, we study the viewpoint dependency in model learning, i.e. the model fails to detect learned categories from complementary viewpoints for both the known and novel classes due to distribution shift. We hypothesize that learning view-invariant features would enhance model robustness to decision-making from diverse viewpoints and address the research question of \textit{learning novel classes from novel viewpoints}. We propose to learn view-invariant spatiotemporal feature embedding for robust action recognition by incorporating adversarial learning to remove view information from the embedded representation and contrastive learning to merge representation from different views of the same activities together. Our contribution of incorporating this complementary information and \textit{k-}face dice statistics into a novel NCD methodology with view-invariant features is summarized below:



(1) We propose a novel joint optimization-based single-stage NCD method NEV-NCD (\textit{N}egative learning, \textit{E}ntropy, and \textit{V}ariance regularization \textit{NCD}) by integrating negative learning, entropy, and variance constraints. Further, we adapt supervised contrastive learning~\cite{khosla2020supervised} to enhance the cluster discrimination.

(2) We validate NEV-NCD in the VAR by experimenting with public UCF101 \cite{soomro2012ucf101} and our in-house multi-view video datasets. 
We thoroughly investigate the impacts of defined class notions, individual loss components, backbone networks, and viewpoints. NEV-NCD achieves $92\%$, $84\%$ test accuracy on the labeled data and $82\%$, $70\%$ average clustering accuracy on the unlabeled data for UCF101 and in-house dataset respectively. Our approaches outperform naive and single-stage NCD baselines in both datasets. 
We open-source our codes and data \footnote{ \href{https://github.com/mxahan/NEV-NCD}{Codes} and \href{https://huggingface.co/datasets/mahmed10/MPSC_MV}{Dataset}} for further research.

(3) We demonstrate the importance of view-invariant feature learning by contrastive and adversarial learning for recognizing novel views for novel actions for robust NCD performance for both novel and known classes from diverse viewpoints by experimenting with a multiview VAR dataset. Our thorough experiments empirically provide insight regarding optimal viewpoints, and data selection settings to develop view-invariant NCD models with computational and labeled efficiency. 


\section{Methodology}

\subsection{Problem Formulation and Notations}

The dataset consists of labeled data $\mathcal{D}^l = \{(x_i^l, y_i)\}_{i = 1}^{N^l}$ where $y_i \in C_L=\{c_1, c_2, ..., c_L\}$ from $L = |C_L|$ unique ordinary categories (instance belongs to the class) and unlabeled data $\mathcal{D}^u = \{(x_i^u)\}_{i = N^l+1}^{N^l + N^u}$ from $U = |C_U|$ latent novel classes $y_i \in C_U=\{c_{L+1}, c_{L+2}, ..., c_{L+U}\}$ with non-overlapping categories $C_L\cap C_U =  \emptyset$. Alternatively, we rewrite $\mathcal{D}^u$ with multiple complementary labels (instance does not belong to complementary label classes) by $\{(x_i^u, \Bar{y}_i)\}_{i = N^l+1}^{N^l+N^u}$, where $x_i^u \notin \Bar{y}_i$ and $\Bar{y}_i \in C_L$. 

Further, the individual class distribution of $\mathcal{D}^l$ and $\mathcal{D}^u$ depends on the captured viewpoint $v_j$ where $j \in {1,2, .. K}$ denotes on one of the $K$ camera view angle positions. The union of these individual view distributions, $\mathcal{D}_{c_i, v_j}$ for $c_j \in C_L\bigcup C_U$, forms the complete data distribution $\mathcal{D}_{c_i}$ for each class as in equation \ref{eq:view_distrib}. 

\begin{equation}
    \mathcal{D}_{c_i} =  \bigcup_{j=1}^K \mathcal{D}_{c_i, v_j}
    \label{eq:view_distrib}
\end{equation}

We denote parameterized 3D convolutional neural network (CNN) for video encoder $f_\theta$, fully connected representation layer $h_\theta$, and fully connected classification head layer $g_\theta$. The $g_\theta$ consists of the concatenation of label head $l_\theta$ ($0$ to $L-1$ labeled categories) and unlabeled head $u_\theta$ ($L$ to $L+U-1$ unlabeled categories) to represent the probabilities over the total dataset categories. The NCD objectives aim to develop DL models to classify the $\mathcal{D}^l$ by comprehending the labeling semantics and clustering the $\mathcal{D}^u$ counterpart based on their underlying latent class distribution.

\subsection{NEV-NCD}

\begin{figure}
 \centering
 \includegraphics[width=\linewidth]{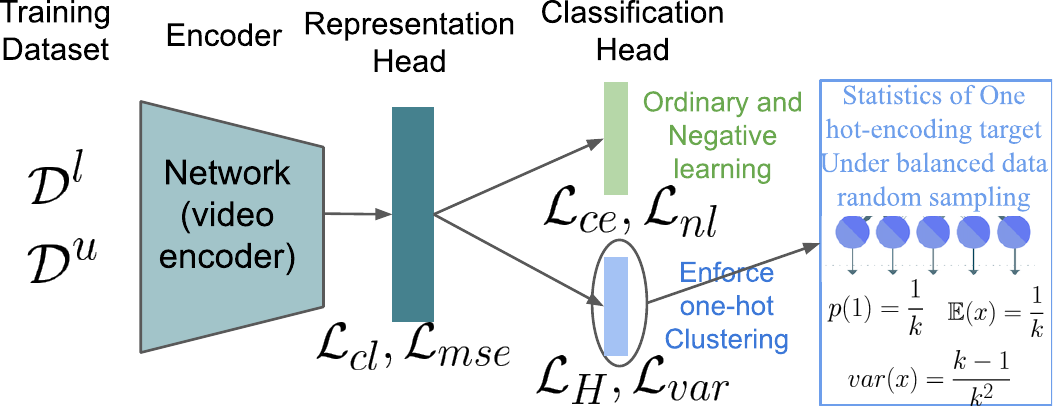}
 \caption{Overview of our NCD approach based on negative learning and variance constraint to identify novel clusters}
 \label{fig:overview_diag}
\end{figure}

Our proposed NEV-NCD method falls under the single-stage approach requiring the availability of both $\mathcal{D}^l$ and $\mathcal{D}^u$ concurrently. NEV-NCD jointly optimizes both the supervised and clustering objectives (Figure \ref{fig:overview_diag}) as discussed in the following.

\textbf{Ordinary Label Learning:}
The supervised training covers learning from both ordinary and multiple complementary labels. The ordinary labeled data $\mathcal{D}^l$ are utilized to optimize the categorical cross-entropy loss \ref{eq:ce_sup} of $y_i \in C_L$ in the classification head. We denote network prediction $\mathbf{\hat{y}_{i}}= \sigma_k(g_\theta(h_\theta (f_\theta (x_i))) )$, where $\sigma_k$ represents Softmax operation over the prediction containing $L+U$ neurons of $g_\theta$. 

\begin{equation}
 \mathcal{L}_{ce} =  - \underset{(x_i, y_i) \in \mathcal{D}^l}{\mathbb{E}}\sum_{j=0}^{L-1} y_{ij} \log{\hat{y}_{ij}} 
 \label{eq:ce_sup}
\end{equation}

$y_{ij}$ represents a one-hot encoding vector of the target $y_i$.

\textbf{Negative Learning:}
We utilize the disjoint condition between $\mathcal{D}^u$ and $\mathcal{D}^l$ latent classes to formulate the problem as learning from multiple complementary labels \cite{feng2020learning} for the $\mathcal{D}^u$ and propose to incorporate negative-learning-based solution \cite{kim2019nlnl}. The negative learning utilizes the complementary labels to reduce their confidence on $l_\theta$ for the corresponding instances contrary to the ordinary labeled supervised learning. In our approach, we enforce zero (low-confidence) for the complementary labels by applying the loss defined in Equation~\ref{eq:neg_learn}.

\begin{equation}
 \mathcal{L}_{nl} =  - \underset{x_i \in \mathcal{D}^u}{\mathbb{E}}[ \sum_{j=0}^{L-1}\log({1-\hat{y}_{ij}})]
 \label{eq:neg_learn}
\end{equation}

\textbf{Entropy Regularization:}
The final classification layer SoftMax operation and negative learning loss function impose higher confidence in the unlabeled classification head $u_\theta$ for $x_i \in \mathcal{D}^u$. We employ the uniquely-labeled data condition and apply entropy constraint loss (Equation~\ref{eq:ent_reg}) to increase the confidence of the unlabeled head as the Dirac-delta distribution minimizes the entropy \cite{grandvalet2004semi}. The entropy regularization further ensures gradient flow through the weight matrix of $u_\theta$ nodes. 


\begin{equation}
 \mathcal{L}_{H} =  - \underset{x_i \in U}{\mathbb{E}}[ \sum_{j=0}^{L+U-1}\hat{y}_{ij}\log{\hat{y}_{ij}}]
 \label{eq:ent_reg}
\end{equation}


\textbf{Contrastive and Consistency Loss:}
We adopt the SupCon \cite{khosla2020supervised} framework with NCD to enhance the representation quality by modifying the data sampling using the underlying data conditions. 
We optimize probabilistic InfoNCE contrastive objectives to learn both categorical and instance discrimination in the representation layer output denoted by $z_i = h_\theta(f_\theta(x_i))$. 

\textit{Category Discrimination:} 
We utilize the known categories of the $\mathcal{D}^l$ and disjoint criterion to perform categorical contrastive learning. Here, we sample query and positive instances from $c_j \in C_L$ and take negative instances distributed from remaining classes and unlabeled data $C_L \slash \{c_j\} \cup C_U$ to minimize the contrastive objective of \ref{eq:l_info} where $s_{i,j} = cosine(z_i, z_j)$, $\tau$ denotes cosine similarity between embedding and temperature.
\begin{equation}
 \mathcal{L}_{cl} = \underset{c_l \sim C_L}{\mathbb{E}}\underset{x_{k,n}| c_k\neq c_l}{\underset{x_i, x_j| c_l}{\mathbb{E}}}-\log\frac{e^{(s_{i,j}/\tau)}}{e^{(s_{i,j}/\tau)} + \sum_{n =0}^N e^{(s_{i,kn}/\tau)}}
 \label{eq:l_info}
\end{equation}

\textit{Instance Discrimination:} Although there lacks information about the intra-class similarity between unlabeled instances, we have their complementary labels. For the query of unlabeled instances, we sample negatives from $\mathcal{D}^l$ to avoid sampling false negative keys and minimize \ref{eq:u_info}. We also enforce consistency \cite{yang2022divide} between the representation of the instance $x_i$ and their augmented version $x_i^\prime$ by equation \ref{eq:mse}. Here, we rely on variance regularization \ref{var_reg} instead of contrastive learning to discriminate between unlabeled latent classes.
\begin{equation}
 \mathcal{L}_{cl} =\underset{x_{k,n}| c_l\in C_L}{\underset{x_i \in \mathcal{D}^u}{\mathbb{E}}}-\log\frac{e^{(s_{i,i^\prime}/\tau)}}{e^{(s_{i,i^\prime}/\tau)} + \sum_{n =0}^N e^{(s_{i,kn}/\tau)}}
 \label{eq:u_info}
\end{equation}


\begin{equation}
 \mathcal{L}_{mse} = \underset{x_i \in \mathcal{D}^u}{\mathbb{E}} ||z - z^\prime||_2^2
 \label{eq:mse}
\end{equation}


\textbf{Variance Regularization:} \label{var_reg} Uniform random sampling from a balanced data of $k$ classes is equivalent to roll a fair $k-$face dice with probabilities, $P = \frac{1}{k}$, of each class. We can define a binary random variable, $X_j \in \{0,1\}$, representing the events of $j-$th face, $j \in {1, 2, .., k}$, outcome success and failure with probabilities, $P(X_j =1) = \frac{1}{k}$, mean, $\mathbb{E}(X_j) = \frac{1}{k}$, and variance $\frac{k-1}{k^2}$.
We explicitly apply this constraint as regularization \ref{eq:var_reg} for the empirical distribution of the sharpened (sharpening hyper-parameter $Sr$) unlabeled prediction head $\Tilde{y}_i =  \sigma_k(\hat{y}_i/Sr)$ for relatively large batch size.

\begin{equation}
 \mathcal{L}_{var} = \underset{x_i \in \mathcal{D}^u}{\mathbb{E}} |var(\Tilde{y}_{i}) - \frac{k-1}{k^2}|_2^2
 \label{eq:var_reg}
\end{equation}








\textbf{Joint Optimization:}
We jointly optimize the learning objective \ref{eq:joint_op} of a weighted ($\lambda$) sum of the discussed losses and regularization by utilizing both $\mathcal{D}^l$ and $\mathcal{D}^u$ in a single stage with supervised contrastive architecture framework. 


\begin{equation}
 \mathcal{L}= \lambda_{ce}\mathcal{L}_{ce} + \lambda_{cl}\mathcal{L}_{cl} + \lambda_{nl}\mathcal{L}_{nl} + \lambda_{H}\mathcal{L}_{H} + \lambda_{var}\mathcal{L}_{var}
 \label{eq:joint_op}
\end{equation}

\textbf{View-invariance constraint:}
We impose optional additional constraints with the joint optimization to learn view-invariant features given the availability of the viewpoint specifications. We adopt adversarial training \cite{tzeng2017adversarial} to enforce the encoder to learn view-invariant discriminative action features by alternatively updating a discriminator and encoder towards an adversarial objective. In our setting, we place a multi-class discriminator $f_d$ to retrieve the source viewpoint information [minimizing the loss of equation \ref{eq:disc_loss}], whereas the encoder network $f_\theta$ aims to learn viewpoint information free action embedding [minimizing the loss of \ref{eq:enc_adv_loss}] to make the $f_d$ task no better than a random guess than maximize the discriminator loss of \ref{eq:disc_loss}.

\begin{equation}
 \mathcal{L}_{d} =  - \underset{\bigcup_{j= 1}^3 c_{i,v_j}}{\mathbb{E}}\sum_{j=0}^{2} \mathbf{e}_j \log{f_d(h_\theta(g_\theta(x_i|c_{i,v_j})))}
 \label{eq:disc_loss}
\end{equation}

Where, $\mathbf{e}_j$ denotes one-hot vector for $j-th$ position. 

\begin{equation}
 \mathcal{L}_{adv} =  - \underset{\bigcup_{j= 1}^3 c_{i,v_j}}{\mathbb{E}}\sum_{j=0}^{2} \mathbf{e}_0 \log{f_d(h_\theta(g_\theta(x_i|c_{i,v_j})))}
 \label{eq:enc_adv_loss}
\end{equation}

Moreover, we enforce contrastive learning objectives to collapse the distributions from different viewpoints to learn a view-invariant representation that only depends on the underlying action class. Particularly, we sample positives by taking class instances from different viewpoints and negatives by sampling instances from other classes following TCN work \cite{sermanet2018time}.




\section{Experiments}

We demonstrate the NEV-NCD effectiveness by experimenting with two video datasets and performing ablation studies.

\subsection{Dataset}

\textbf{UCF101 \cite{soomro2012ucf101}:} We have preprocessed the UCF101 data with randomly selected $91$ labeled actions and the remaining $10$ as unlabeled actions. We balance the distribution of the instances from these unlabeled classes to match the NCD assumption.

\textbf{In-house dataset:} Our collected multiview dataset contains three synchronous views $K=3$ from three cameras (view 0, $v_0$: horizontally positioned wide-lense static action camera, view 1, $v_1$:horizontally positioned handheld smartphone and view 2, $v_2$: low-altitude flying drone camera) that captured the action from three realistic positions with varying camera settings, positions, and angles. 
These camera positions are designed to capture complementary views with respect to each other while the volunteers perform actions without any directional preferences. 
It contains class-balanced ten regular micro-actions with both static (i.e. body poses: sitting, standing, lying with face up and down) and dynamic patterns (i.e. temporal patterns: walking, push up, waving hand, leg exercise, object carrying, and pick/drop). In total, we collect $\approx~14$ hours of video from $11$ volunteers under varying realistic lighting, environments, and backgrounds.
We perform controlled experiments to study various NEV-NCD components, view-invariant, and action selection using our in-house dataset due to its scalable label space and available viewpoint meta-data.

\subsection{Implementation Details}
\textbf{Network Architecture:} We experimented with three different CNN architectures (SlowFast, R2+1D \cite{tran2015learning}, and P3D \cite{qiu2017learning}) that have demonstrated their capabilities in VAR as encoder design choices. We experimented with both random initialization and weight transfer from the kinetics dataset pre-trained model weights available in PyTorch Modelzoo and remove their classification head. We placed a $1000-$dimensional embedding layer with ReLU activation and classification head with SoftMax activation on top of encoder features.

\textbf{Losses:} Inspired by \cite{liu2022residual}, we applied adaptive weights for the contrastive learning, negative learning, variance loss, and cross-entropy loss components with the epoch number, $n_{ep}$, throughout the learning phase as shown in equation \ref{eq:var_weight} and \ref{eq:ce_weight}. In our experimentations, We fixed entropy loss weights ($\lambda_{H} = 1.0$) and other hyper-parameters ($\tau = 0.05$, $Sr = 0.1$). 
To calculate the empirical distribution variance, we traded off between computational cost and the law of large numbers and chose the batch size of $10 \times U$ instances. 

\begin{equation}
     \lambda_{cl} = \lambda_{nl} =  \lambda_{var} = 0.2 +0.5 \times n_{ep} 
     \label{eq:var_weight}
\end{equation}
\begin{equation}
        \lambda_{ce} =  \max (0, 1-0.01 \times n_{ep}) + 0.5
        \label{eq:ce_weight}
\end{equation}

\subsection{View experiments with NCD}

We perform extensive experiments [table \ref{table:view_main} and \ref{table:view_ablate}] in multiple settings to demonstrate the importance of viewpoint and view-invariant feature learning by providing viewpoint information and incorporating view-invariant constraints in the NCD training phase for both labeled and unlabeled data. Firstly, we consider all three viewpoints for all the labeled classes while considering only single viewpoint instances in the unlabeled data and vice versa. 
secondly, we incorporate the partial viewpoints for the labeled and unlabeled classes by obfuscating one of the views in the training phases. 
Thirdly, we study the optimal viewpoints by providing viewpoints only for selective classes. 
Further, we also develop models by ignoring view-invariant constraints in all experiments for baseline comparison.








\subsection{Baseline Comparison and Ablation Studies}

We compare our methodology with three naive approaches -- (i) the supervised training using only the labeled parts, (ii) instance discrimination-based contrastive learning utilizing a complete dataset, and (iii) the SupCon approach combining both labeled and unlabeled data. Additionally, we compare the performance of NEV-NCD with the most relevant single-stage joint optimization-based state-of-the-art NCD method UNO~\cite{fini2021unified} with single head Sinkhorn-Knopp clustering (replacing the swapped prediction with self-labeling and retraining \cite{asano2019self}) by adapting it for our VAR scenario.


We run numerous experiments to study the impacts of different design choices toward better understanding the proposed NCD VAR system.
Firstly, we investigate the impacts and the relative importance of each loss component of our proposed joint loss. We also vary the hyper-parameters of the loss components ($\lambda_{nl}$, $\lambda_{nl}$, negative sample number) to study their robustness. 
Secondly, we investigate the impacts of choosing different 3D CNN architectures (i.e. P3D, R2+1D, SlowFast) as the parameterized feature encoders for our VAR model.
Finally, we inspect the effect of multiview properties by varying members of $C_L, C_U$, and multiple viewpoint variations for the actions using our in-house dataset.




\section{Results}

We evaluated NEV-NCD in left-out labeled and unlabeled test clips by both quantitative and qualitative metrics for both datasets. 
To scrutinize the quality (i.e. separability of classes) of the learned embeddings, we employed low-dimensional t-SNE visualization.
We quantified the performance over the labeled and unlabeled data by deploying standard top-1 accuracy (ACR) and average clustering accuracy (ACC) (equation \ref{eq:acc_mat}), respectively. 
\begin{equation}
 ACC = \underset{p \in Pe}{\max} \frac{1}{N} \sum_{i = 1}^N \mathbbm{1} \{y_i = p({\hat{y}_i})\}
 \label{eq:acc_mat}
\end{equation}
\noindent Here $y_i \in C_U$ and $\hat{y}_i$ represent the ground-truth and cluster prediction of a sample $x_i \in \mathcal{D}^u$, respectively. $Pe$ denotes the set of all permutations and can be efficiently computed $\forall x_i$ \cite{kuhn1955hungarian}.

\subsection{Main Results and comparisons}

\begin{table}[!htb]
\centering
\footnotesize
\caption{Summarized Result on UCF101 Dataset (90 labeled, R2+1d)}
\label{table:result_ucf101}
\begin{tabular}{|c|c|c|c|l|}
\hline
Methods  & NEV-NCD & UNO  & SpCn & Sup \\ \hline
Accc. \% & 92.4    & 92.1 & 93.2 & 92.6      \\ \hline
ACC \%   & 82.7    & 79.8 & 24.1 & 16.4      \\ \hline
\end{tabular}
\end{table}

\begin{table}[!htb]
\centering
\footnotesize
\caption{Overall Result on our in-house video dataset}
\label{table:result_all}
\begin{tabular}{|c|ccc|c|cc|}
\hline
& \multicolumn{3}{c|}{Loss Components} &  & \multicolumn{2}{c|}{Metrics}  \\ \cline{2-4} \cline{6-7} 
\multirow{-2}{*}{Methods}  & \multicolumn{1}{c|}{\begin{tabular}[c]{@{}c@{}}$\mathcal{L}_{ce}$, \\ $\mathcal{L}_{cl}$\end{tabular}} & \multicolumn{1}{c|}{SK} & \begin{tabular}[c]{@{}c@{}}$\mathcal{L}_{nl}$, \\ $\mathcal{L}_{var}$, \\ $\mathcal{L}_{H}$\end{tabular} & \multirow{-2}{*}{\begin{tabular}[c]{@{}c@{}}Labeled \\ Classes\end{tabular}} & \multicolumn{1}{c|}{\begin{tabular}[c]{@{}c@{}}ACR\\ (\%)\end{tabular}}  & \begin{tabular}[c]{@{}c@{}}ACC\\ (\%)\end{tabular}\\ \hline\hline
\begin{tabular}[c]{@{}c@{}}Ours\\ 
(R2+1d)\end{tabular}& \multicolumn{1}{c|}{Y} & \multicolumn{1}{c|}{N}& Y  & \begin{tabular}[c]{@{}c@{}}Uniform\\ Dynamics\\ Static\end{tabular} & \multicolumn{1}{c|}{\begin{tabular}[c]{@{}c@{}}82.9\\ 81.8\\ 87.1\end{tabular}} & \begin{tabular}[c]{@{}c@{}}67.3\\ 64.2\\ 53.0\end{tabular} \\ \hline
\begin{tabular}[c]{@{}c@{}}Ours\\ (SlowFast)\end{tabular}& \multicolumn{1}{c|}{Y} & \multicolumn{1}{c|}{N}& Y  & \begin{tabular}[c]{@{}c@{}}Uniform\\ Dynamics\\ Static\end{tabular} & \multicolumn{1}{c|}{\begin{tabular}[c]{@{}c@{}}84.1\\ 82.7\\ 87.8\end{tabular}} & \begin{tabular}[c]{@{}c@{}}70.5\\ 68.2\\ 53.2\end{tabular} \\ \hline
\begin{tabular}[c]{@{}c@{}}Ours\\ (P3D)\end{tabular}  & \multicolumn{1}{c|}{Y} & \multicolumn{1}{c|}{N}& Y  & \begin{tabular}[c]{@{}c@{}}Uniform\\ Dynamics\\ Static\end{tabular} & \multicolumn{1}{c|}{\begin{tabular}[c]{@{}c@{}}76.5\\ 72.9\\ 78.6\end{tabular}} & \begin{tabular}[c]{@{}c@{}}62.3\\ 59.3\\ 45.7\end{tabular} \\ \hline
Sup & \multicolumn{1}{c|}{$\mathcal{L}_{ce}$}& \multicolumn{1}{c|}{N}& N  & Uniform & \multicolumn{1}{c|}{84.6} & 12.5 \\ \hline

UNO \cite{fini2021unified} & \multicolumn{1}{c|}{$\mathcal{L}_{ce}$}& \multicolumn{1}{c|}{Y}& N  & Uniform & \multicolumn{1}{c|}{83.4} & 62.1 \\ \hline

SpCn \cite{khosla2020supervised} & \multicolumn{1}{c|}{Y} & \multicolumn{1}{c|}{N}& N  & Uniform & \multicolumn{1}{c|}{85.2} & 12.8 \\ \hline
CL \cite{chen2020simple} & \multicolumn{3}{c|}{Fine tuned}& Uniform & \multicolumn{1}{c|}{80.6} & 13.4 \\ \hline
& \multicolumn{1}{c|}{Y} & \multicolumn{1}{c|}{N}& $\mathcal{L}_{nl}$& Uniform & \multicolumn{1}{c|}{83.9} & 25.3 \\ \cline{2-7} 
& \multicolumn{1}{c|}{Y} & \multicolumn{1}{c|}{N}& $\mathcal{L}_{var}$  & Uniform & \multicolumn{1}{c|}{85.1} & 15.2 \\ \cline{2-7} 
& \multicolumn{1}{c|}{Y} & \multicolumn{1}{c|}{N}& $\mathcal{L}_H$ & Uniform & \multicolumn{1}{c|}{84.8} & 25.6 \\ \cline{2-7} 
& \multicolumn{1}{c|}{Y} & \multicolumn{1}{c|}{N}& $\mathcal{L}_{nl}$, $\mathcal{L}_{H}$  & Uniform & \multicolumn{1}{c|}{83.9} & 30.2 \\ \cline{2-7} 
\multirow{-5}{*}{\begin{tabular}[c]{@{}c@{}}Ablation \\ Studies\end{tabular}} & \multicolumn{1}{c|}{Y} & \multicolumn{1}{c|}{N}& $\mathcal{L}_{nl}$, $\mathcal{L}_{var}$& Uniform & \multicolumn{1}{c|}{83.7} & 60.5 \\ \hline
\end{tabular}
\end{table}

We report quantitative results over the left-out test data of the UCF101 and the in-house dataset in Table ~\ref{table:result_ucf101} and ~\ref{table:result_all} respectively. In both cases, NEV-NCD outperforms others in clustering the $\mathcal{D}^u$ (high ACC) while classifying the $\mathcal{D}^l$ (high ACR).

\begin{figure}
    \centering
    \includegraphics[width = \linewidth]{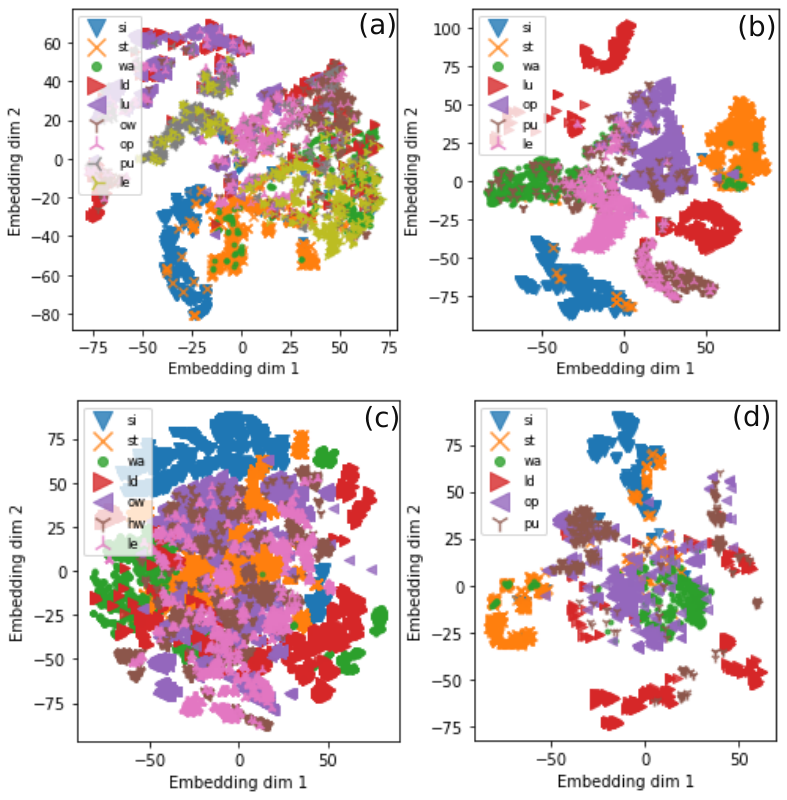}
    \caption{2D t-SNE plot for the representation of the embedding layer (a) UNO (b) NEV-NCD (c) Supervised Baseline (d) Negative Learning}
    \label{fig:t-sne-comp}
\end{figure}

The naive baselines fail to cluster the $\mathcal{D}^u$ separately (low ACC) from the labeled data embedding without external constraints. NEV-NCD outperforms the comparable UNO (self-labeling via Sinkhorn-Knopp (SK) clustering and retraining on pseudo-label) in the ACC metrics without retraining. Simultaneously, we analyze the 2D t-SNE plot for the embedding layer output. As depicted in figure \ref{fig:t-sne-comp}, we observe that NEV-NCD learns distinguished features for both labeled and unlabeled instances.

The NCD performance drops significantly in the absence of $\mathcal{L}_{var}$ and $\mathcal{L}_{nl}$ as reported in Table~\ref{table:result_all}. The $\mathcal{L}_{nl}$ enforces a boundary between labeled and unlabeled classes by pushing away the unlabeled embedding from the labeled embedding. However, the $\mathcal{L}_{nl}$ results in collapsing unlabeled classes into a single cluster without additional constraints. The $\mathcal{L}_{var}$ provides the equipartition cluster constraint and avoids trivial collapses for the embedding of unlabeled instances. The $\mathcal{L}_{H}$ loss increase inter-class separation by enforcing unique decision for the unlabeled classification heads. Additionally, we study and report the impact of relative weights of the loss components in Table \ref{table:result_ablation}. 

We also observe the impact of labeled class notions in unlabeled data clustering. In our VAR experiments, labeled class sets with only static categories (sitting, standing, etc.) reduce the performance of clustering of dynamic actions (walking, push-ups, etc.). Besides, we find that NCD methods learn the invariant representations of actions from different viewpoints of unlabeled data by providing multiview labeled instances. Finally, we notice that the strong function class improves the overall performance in both labeled and unlabeled data (Table~\ref{table:result_all}).

\begin{table}[]
\centering
\footnotesize
\caption{Ablation Studies using in-house dataset for different constant $\lambda_{nl}$ and $\lambda_{var}$ with SlowFast model (fixed other parameters).}
\label{table:result_ablation}
\begin{tabular}{ccccl}
\hline
\multicolumn{2}{c}{Weights}      & \multicolumn{2}{c}{Metrics} & Comment       \\ \hline
$\lambda_{nl}$ & $\lambda_{var}$ & ACR (\%)    & ACC (\%)    &               \\
0.1 $\sim$ 1        & 1               & 84            & 65 $\sim$ 70       & \multirow{2}{*}{\begin{tabular}[c]{@{}l@{}} Slower convergence \\for smaller values\end{tabular}} \\
1              & 0.1 $\sim$ 1           & 84            & 64 $\sim$ 70       &               \\ \cline{1-5}
\end{tabular}
\end{table}

\subsection{View-point Ablation Results}

We report our results over the left-out-test person data from all three views $v_{0,1,2}$ for all the actions in table \ref{table:view_main}. We quantitatively analyze the embedding clustering properties by measuring the silhouette coefficient (SC). We also measure the classification performance for the labeled and unlabeled classes by reporting the accuracy and average clustering accuracy (ACC) metrics. In our experimentation, we consistently observe the improved clustering performance by adding view-invariance (VI) constraints. 

We also empirically study the impact of individual views and report our findings in Table~\ref{table:view_ablate}. We find the most complementary viewpoints provide the optimal data instances to develop the robust model. In our in-house data, $v_0$ and $v_1$ share common viewpoints as they capture horizontal views, and $v_2$ provides alternative vertical observations concerning the other two. We observe a performance boost for matching data quantity by including complementary viewpoints by combining $v_0$ or $v_1$ with $v_2$. 





\begin{table}[]
\centering
\footnotesize
\caption{Results for view-invariant (VI) losses for NEV-NCD setting with R$2+1$d model over the in-house dataset.}
\label{table:view_main}
\begin{tabular}{|l|ll|l|l|l|l|}
\hline
\multirow{2}{*}{\begin{tabular}[c]{@{}l@{}}Exp\\ No\end{tabular}} & \multicolumn{2}{l|}{Train Data} & \multirow{2}{*}{\begin{tabular}[c]{@{}l@{}}VI\\ Loss\end{tabular}} & \multirow{2}{*}{SC} & \multirow{2}{*}{Accu.} & \multirow{2}{*}{ACC} \\ \cline{2-3}
     & \multicolumn{1}{l|}{L}    & U   &  &                     &     &   \\ \hline
1    & \multicolumn{1}{l|}{  $v_{0,1,2}$}  &   $v_{0,1,2}$ & N     & 0.51                & 82.9                   & 67.3                 \\ \hline
2    & \multicolumn{1}{l|}{  $v_{0,1,2}$}  &   $v_{0,1,2}$ & Y     & 0.56                & 83.3                   & 67.5                 \\ \hline
3    & \multicolumn{1}{l|}{  $v_{0,1,2}$}  & $v_1$   & N     & 0.38                & 83.2                   & 49.3                 \\ \hline
4    & \multicolumn{1}{l|}{  $v_{0,1,2}$}  &  $v_1$ & Y     & 0.51               & 83.1                   & 62.8                 \\ \hline
5    & \multicolumn{1}{l|}{$v_1$}    &   $v_{0,1,2}$ & N     & 0.31               & 53.5                   & 62.1                 \\ \hline
6    & \multicolumn{1}{l|}{$v_1$}    &   $v_{0,1,2}$ & Y     & 0.45                & 76.3                   & 66.7                 \\ \hline
\end{tabular}
\end{table}


\begin{table}[]
\centering
\footnotesize
\caption{Results for viewpoint and view-invariant loss importance for NEV-NCD setting with R$2+1$d model.}
\label{table:view_ablate}
\begin{tabular}{|l|ll|l|l|l|l|}
\hline
\multirow{2}{*}{\begin{tabular}[c]{@{}l@{}}Exp\\ No\end{tabular}} & \multicolumn{2}{l|}{Train Data} & \multirow{2}{*}{\begin{tabular}[c]{@{}l@{}}VI\\ Loss\end{tabular}} & \multirow{2}{*}{SC} & \multirow{2}{*}{Accu.} & \multirow{2}{*}{ACC} \\ \cline{2-3}
        & \multicolumn{1}{l|}{L}    & U   &          &                     &                        &                      \\ \hline
1       & \multicolumn{1}{l|}{  $v_1, v_2$}  &   $v_1, v_2$ & N        & 0.19                & 65.6                   & 51.5                 \\ \hline
2       & \multicolumn{1}{l|}{  $v_1, v_2$}  &   $v_1, v_2$ & Y        & 0.36                & 71.2                   & 56.7                 \\ \hline
3       & \multicolumn{1}{l|}{  $v_0, v_1$}  &   $v_0, v_1$ & N        & 0.12                & 54.4                   & 43.1                 \\ \hline
4       & \multicolumn{1}{l|}{  $v_0, v_1$}  &   $v_0, v_1$ & Y        & 0.23                & 61.8                   & 47.6                 \\ \hline
\end{tabular}
\end{table}

\section{Discussion and future works}
Our current method leverages three typical NCD data assumptions: (i) prior knowledge about the number of latent classes for the unlabeled data (unknown target cluster quantity), (ii) disjoint set condition between the $C_L$ and $C_U$ (negative learning), and (iii) the uniform distribution among unlabeled classes (variance regularization). Moreover, the empirical variance estimation requires a large batch size relative to $U$ randomly sampled from unlabeled data as per the strong law of large numbers. 

In the future, we aim to challenge the NCD assumptions and investigate the minimal constraint required for the unlabeled data to solve the NCD clustering. We plan to extend our works towards generalized categories discovery \cite{vaze2022generalized} and explore over-clustering, self-training, and distribution aligning solutions. We also plan to integrate and investigate the impact of attention-based learners (vision transformers) in NCD VAR applications.

Due to the combinatorial possibilities of viewpoint experiments, we report our results to a plausible set of experiments to demonstrate the importance of multiview data and view-invariant constraints for robust model development. 
However, the optional view-invariant constraint increases the NCD model's robustness against viewpoint variation with an expanse of additional viewpoint information associated with the instances and the computational cost of the adversarial components. Alternatively, synchronous video data can alleviate such requirements as it enables to implementation of temporal contrastive learning TCN \cite{sermanet2018time} and TCLR \cite{dave2022tclr} to learn view-invariant representation.

\section{Conclusion}
In this work, we introduced and studied the negative learning, entropy, and variance regularization in NCD research to propose a novel single-stage NCD method NEV-NCD removing additional pseudo-label generation and retraining steps. We validated NEV-NCD by performing extensive experiments with the VAR domain using 3D CNN-based networks. 
The qualitative and quantitative results demonstrate the impacts of NEV-NCD components and compatibility with state-of-the-art NCD works. We also demonstrate that the NCD losses alone fail to capture novel categories from novel viewpoints. Finally, we propose and validate that adding view-invariant constraint improves model robustness and enhances models' capacity to identify novel categories from novel viewpoints.



\bibliographystyle{IEEEbib}
\bibliography{refs}

\end{document}